\documentclass{article}

\usepackage{arxiv}

\usepackage[utf8]{inputenc} 
\usepackage[T1]{fontenc}    
\usepackage{hyperref}       
\usepackage{url}            
\usepackage{booktabs}       
\usepackage{amsfonts}       
\usepackage{nicefrac}       
\usepackage{microtype}      
\usepackage{lipsum}
\usepackage{algorithm}
\usepackage{algorithmic}
\usepackage[T1]{fontenc}
\usepackage{graphicx}
\usepackage{caption}
\usepackage{subcaption}

\usepackage{amsmath}
\usepackage{booktabs}

\title{Novelty Detection and Learning with Extremely Weak Supervision}

\author{
  Eduardo Soares \\
  School of Computing and Communications\\
 LIRA Research Centre\\
 Lancaster University\\
  Lancaster, LA1 4WA, UK \\
 \texttt{e.almeidasoares@lancaster.ac.uk} \\
   \And
 Plamen Angelov \thanks{Honorary Professor, Technical University, Sofia, Bulgaria.} \\
 School of Computing and Communications\\
  LIRA Research Centre\\
  Lancaster University\\
  Lancaster, LA1 4WA, UK \\
  \texttt{p.angelov@lancaster.ac.uk} \\
}

\begin{document}
\maketitle

\begin{abstract}
In this paper we offer a method and algorithm, which make possible fully autonomous (unsupervised) detection of new classes, and learning following a very parsimonious training priming (few labeled data samples only). Moreover, new unknown classes may appear at a later stage and the proposed xClass method and algorithm are able to successfully discover this and learn from the data autonomously. Furthermore, the features (inputs to the classifier) are automatically sub-selected by the algorithm based on the accumulated data density per feature per class. As a result, a highly efficient, lean, human-understandable, autonomously self-learning model (which only needs an extremely parsimonious priming) emerges from the data. To validate our proposal we tested it on two challenging problems, including imbalanced Caltech-101 data set and iRoads dataset. Not only we achieved higher precision, but, more significantly, we only used a single class beforehand, while other methods used all the available classes) and we generated interpretable models with smaller number of features used, through extremely weak and weak supervision. 

\end{abstract}


\section{Introduction}
\label{intro}
Machine learning and pattern recognition are at the peak of their development with sharp interest not only from scientists and practitioners, but also from the wider public and media. This is, in part, thanks to the boom surrounding the wider area of artificial intelligence (AI) and recent successful and widely publicized applications ranging from driverless cars \cite{waldrop2015autonomous,chen2015deepdriving}, defense and security \cite{allen2017artificial,tyugu2011artificial,xu2016toward}, to home applications \cite{robles2010applications,shah2016survey}. 

Despite the great success, the underlying concept of machine learning which requires large amount of labeled training data remains unchanged. Moreover, the most powerful state-of-the-art approaches such as deep neural networks and SVM suffer from lack of interpretability \cite{rudin2019stop,angelov2018toward,core2006building}. Moreover, are extremely power-, time- and computational- resources hungry and unable to adapt and change with agility. They require complete retraining even for a single or few new data samples.

In this paper we propose a method and algorithm that departs from the traditional approach and offers a paradigm shift bringing the machine learning, in general, and pattern recognition and classification, in particular, extremely close to a fully unsupervised form. It offers a self-learning locally generative models that work together and require extremely light supervision in the form of few data samples. It is able to automatically detect the unknown and to learn from it. This is in sharp contrast to the traditional approach where learning is, in essence, only an averaging of the history. The current approaches struggle to detect changes, dynamical evolution or appearance of new classes. They also assume a certain number of features (the same for all classes) provided at the start of the process. This is one of the reasons traditional approaches struggle to predict or react quickly to sudden changes in the data pattern, such as the economic crash during 2008 \cite{hodgson2009great}, for example.

The proposed approach is based on using prototypes and learning locally around them extracting the empirical data distribution called typicality as well as the data density \cite{angelov2019empirical}. The approach is recursive, thus computationally very lean. It is also non-iterative, non-parametric. This adds to its efficiency in terms of time and computational resources. From the user perspective, the proposed approach is clearly understandable to human users since it can be represented in a linguistic $IF ... THEN$ form. It combines reasoning and logic with machine learning. It can also be presented as a deep neural network. Finally, it also has a statistical nature and offers an empirical form of the pdf \cite{angelov2017generalized}.

In this paper we apply this new principally different type of machine learning to four challenging problems and demonstrate its significant advantages.

The remainder of this paper is organized as follows: The concept and algorithm section introduces the proposed exploratory approach for extremely weakly supervised classification. The experimental data employed in the analysis and results are presented in the results section. Conclusion is presented in the last section of this paper.

\section{Concept and Basic Algorithm}

The pipeline of learning from data includes the following steps:

1)\textbf{Pre-precessing}, which includes different substeps like normalization/standardization, dealing with missing data, and feature selection \cite{kotsiantis2006data}. Specifically for image processing there are often other stages, such as rotation, augmentation, scaling, elastic deformation, etc \cite{russ2016image}. Even deep learning methods which claims to avoid handcrafting apply some of the cited steps.

2)\textbf{Learning phase}, which can be offline, when the complete dataset is available; or it can be done online, when the data arrive in the form of a data stream (sample-by-sample). Evolving learning, ability of the algorithms to adapt their parameters and structure according to the non-stationary data streams, is a more sophisticated form of online learning \cite{angelov2010evolving,vskrjanc2019evolving}.

3)\textbf{Generating outputs} for new unseen data, which is the \textbf{validation phase}. Different algorithms use different strategies in order to validate the model generated in the learning phase.

The proposed method also starts with a pre-processing step which involves mostly the same steps depending on the specific problem. 

The proposed xClass method uses standardization and normalization as follows:

Firstly,it standardize the newly observed data sample, $x_{i}$; where $i=1,2,...,n$ denotes a time stamp in the current moment. $j=1,2,...,n$ refers to the number of features of the given $x$.

\begin{eqnarray}
\widehat{x}_{i,j} = \frac{x_{i,j} - \mu(x_{i,j})}{\sigma(x_{i,j})}
\label{Si}
\end{eqnarray}

\noindent where $\widehat{x}$ denotes the standardized data sample. Outliers $(|\widehat{x}|\geq 3)$ are ignored and not used for training.
After that, the data is rescaled within the range $[0,1]$ to consider them in the same proportion. It is important to highlight that in the proposed xClass method, the normalization is done upon the standardized data. Unity-based normalization of the $i$-th element of the $j$-th sample is given by:

\begin{eqnarray}
\Bar{x}_{i,j} = \frac{\widehat{x}_{i,j} - \min\limits_{i}(\widehat{x}_{i,j})}{\max\limits_{i}(\widehat{x}_{i,j})-\min\limits_{i}(\widehat{x}_{i,j})}
\label{Si}
\end{eqnarray}

\noindent where $\Bar{x}$ denotes the normalized data sample.

The prototype-based learning is the core of the proposed method which represents local (the prototypes are focal points of locally valid generative models described by multimodal Cauchy distribution \cite{angelov2019empirical}. The meta-parameters are initialized with the first observed data sample. The proposed algorithm works per class; therefore, all the calculations are done for each class separately. 

\begin{equation}
P \leftarrow 1;~~~\mu \leftarrow \Bar{x}_{i};
\end{equation}
where $\mu$ denotes the global mean of data samples of the given class. $P$ is the number of the identified prototypes in total from the observed data samples.

Each class $C$ is initialized by the first data sample of that class:
\begin{equation}
\begin{split}
\mathrm{C}_{1} \leftarrow \Bar{x}_{1};~~~p_{1} \leftarrow \Bar{x}_{1};\\
S_{1} \leftarrow 1;~~~r_{1}\leftarrow r^*;
\end{split}
\end{equation}

where, $p_{1}$ is the prototype of $\mathrm{C}_{1}$; $S_{1}$ is the corresponding support (number of members); $r_{1}$ is the corresponding radius of the area of influence of $\mathrm{C}_{1}$.

In this paper, we use $r^*=\sqrt{2-2cos(30^o)}$ same as \cite{angelov2019empirical}; the rationale is that two vectors for which the angle between them is less than $\pi/6$ or $30^o$ are pointing in close/similar directions. That is, we consider that two feature vectors can be considered to be similar if the angle between them is smaller than 30 degrees. Note that $r^*$ is data derived, not a problem- or user- specific parameter. In fact, it can be defined without \textit{prior} knowledge of the specific problem or data. The next step is to calculate the data density at $\Bar{x}_{i}$ and $p_{j}~(j=1,2,...,P)$.

\begin{equation}
 D^f_{i}(\Bar{x}^f_{i})=\frac{1}{1+\frac{ (\Bar{x}^f_{i}-p_{i}^f)}{1-\mu_{i}^f}}.
\label{eqy}
\end{equation}

The reason it is Cauchy is not arbitrary \cite{angelov2019empirical}. It can be demonstrated theoretically that if Euclidean or Mahalanobis type of distances in the feature space are considered, the data density reduces to Cauchy type as referred in equation (\ref{eqy}). It can also be demonstrated that the so called \textit{typicality}, $\tau$, which is the weighted average of the data density, $D$, with weights representing the frequency of occurrence of a data sample \cite{angelov2019empirical}. Furthermore, the \textit{typicality}, $\tau$ can be considered an empirically derived form of the pdf having the same properties, notably, it integrates to $1$ an infinite range. 

Density per feature $f$ is obtained according to the equation (\ref{eqy}), where $D^f_{i}$ denotes the density for $f$-th feature of the $\Bar{x}_{i}$ sample. 

The cumulative effect across all data samples per feature can be obtained according to the equation (\ref{x}).

\begin{eqnarray}
\Lambda_{i}^f = \frac{\Sigma_{i=1}^{n} D^f_{i}(\Bar{x}^f_{i})}{n}.
\label{x}
\end{eqnarray}

 The cumulative contribution for each feature $\Lambda^f_{i}$ can be rank ordered, $n$ represents the number of samples. The higher, the value of $\Lambda_{i}^f$ is for a particular feature, the more important is the $f$-th feature. The rationale is that an interesting feature has higher density than other features - meaning that it conveys unique, different clear information, and, as a consequence, it contributes more to the classifier's result because the overlap between data of different classes is less pronounced for this feature. 

Then the algorithm absorbs the new data samples one by one by assigning then to the nearest (in the feature space) prototype:

\begin{equation}
n^*=\operatorname*{argmin}_{j=1,2,...,P}(||\Bar{x}_{i}-p_{j}||^2)
\end{equation}

Because of this form of assignment, the shape of the data partitioning is of the so-called Voronoi tesselation type \cite{okabe2009spatial}. We call all data points associated with a prototype \textit{data clouds}, because their shape is not regular (e.g., hyper-spherical, hyper-ellipsoidal, etc.) and the prototype is not necessarily the statistical and geometric mean \cite{angelov2019empirical}. 

In case, the following condition \cite{angelov2019empirical} is met:
\begin{equation}
\begin{split}
\textit{\text{IF }}(D_{i}(\Bar{x}_{i})\geq\max_{j=1,2,...,P}D_{i}(p_j))~~
\textit{\text{OR }}~~(D_{i}(\Bar{x}_{i})\leq \min_{j=1,2,...,P}D_{i}(p_j))\\
\textit{\text{ THEN }} (add~a~new~data~cloud)\label{eq8}
\end{split}
\end{equation}

It means that $\Bar{x}_{i}$ is out of the influence area of $p_j$. Therefore, $\Bar{x}_{i}$ becomes a new prototype of a new \textit{data cloud} with meta-parameters initialized by equation (\ref{eqIN}).
 Add a new data cloud:
\begin{equation}
\begin{split}
P \leftarrow P+1;~~~\mathrm{C}_{P} \leftarrow \{\Bar{x}_{i}\};
p_{P} \leftarrow \Bar{x}_{i};~~~S_{P} \leftarrow 1;
~r_{P}\leftarrow r_o;
\end{split}
\label{eqIN}
\end{equation}

Otherwise, data cloud parameters are updated online by equation (\ref{update}). It has to be stressed that all calculations per data cloud are performed on the basis of data points associated with a certain data cloud only (i. e. locally, not globally, on the basis of all data points).

\begin{equation}
\begin{split}
\mathrm{C}_{n^*} \leftarrow \mathrm{C}_{n^*} +\Bar{x}_{i}; ~~
p_{n^*} \leftarrow \frac{S_{n^*}}{S_{n^*}+1}p_{n^*}+\frac{S_{n^*}}{S_{n^*}+1}Bar{x}_{i}; ~~ \\
S_{n^*} \leftarrow S_{n^*}+1; ~~
r_{n^*}^2 \leftarrow \frac{r_{n^*}^2 +(1-||p_{n^*}||^2)}{2};
\end{split}
\label{update}
\end{equation}

One of the strongest aspects of the proposed approach is its high level of interpretability which comes from its prototype-based, local generative models as well as as its ability to be expressed as a set of linguistic $IF ... THEN$ fuzzy rules of the following type:

$R:IF (x \sim p_1)~~OR ~~ ... ~~OR~~(x\sim p_{P}) ~THEN~(Class~c)$

The fuzziness represents the degree of association/similarity to the prototypes. Indeed, the value of data density, $D$, equation (\ref{eqy}) can be interpreted as a membership function of the fuzzy set $(x \sim p)$ \cite{angelov2019empirical}. With a maximum $1$ when $x=p$. The continuous typicality, $\tau$ given by the equation (\ref{tau}), is an empirically derived form of probability distribution.
The value of $\tau$ even at the point $x=p_i$ is much less than 1 the integral of $\int_{-\infty }^{\infty}\tau dx =1$.
\begin{equation}
\tau_i(x)= \frac{D_i(x)}{\int _x D_i(x)dx}
\label{tau}
\end{equation}
\subsection*{Detect \& Learn from Unknown}
This is the most innovative part of the proposed algorithm in addition to the feature selection per class, which wakes it exploratory (we call it xClass) and allows to detect new data patterns autonomously and learn from them.

\subsubsection*{Drop of confidence (detect the novelty)}
Unlabeled data samples become available as soon as the training process with labeled samples finishes. Then, the eXploratory classifier (xClass) can continue to learn from these unknown data samples. The unlabeled training samples are defined as the set $\left \{ u \right \}$, and the number of unlabeled samples is defined as $U$. 

The first step in the weakly supervised learning process of xClass, is to extract the vector of confidence/degrees of closeness to the nearest prototypes for each unlabeled data sample defined as $\lambda(u_i)$, $i=1,2,...,U$ follows: 

\begin{eqnarray}
\lambda = \underset{j=1,2,...,P}{max} (\Bar{x},p_j),
\label{eq1}
\end{eqnarray}

\noindent where $\lambda$ denotes the confidence degree. 

The recursive mean $\overline{\mu}_i$ of the $\lambda^{max}$ for the labeled data samples is used to detect sudden drop of the confidence generated by the xClass classifier when a new unknown class arrives and can be calculated as follows \cite{angelov2012autonomous}:

\begin{equation}
\overline{\mu_i} = \frac{i-1}{i}\overline{\mu}_{i-1}+\frac{1}{i}\lambda^{max}_i, \overline{\mu}_1=\lambda^{max}_1.
\label{recursive}
\end{equation}

\begin{figure}
\centering
\includegraphics[width=12cm]{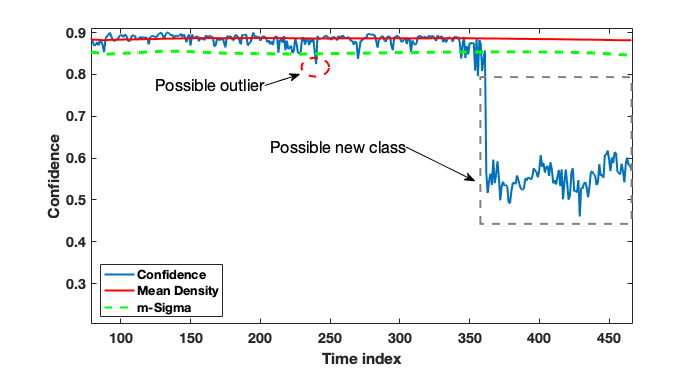}
\caption{Drop of confidence of the proposed method when a new a unseen class arrives}
\label{fig:fall}
\end{figure}

Then the m-$\sigma$ rule is applied, for detailed explanation about the m-$\sigma$ please refer to \cite{8851842}. New classes are actively added by the proposed xClass classifier when the inequality (\ref{eq3}) is satisfied and rules are actively created. Otherwise, if the inequality is not satisfied the newly arrival unlabeled data samples are used for updating the structure and meta-parameters of the xClass classifier. Fig. \ref{fig:fall} illustrates the drop of confidence of the proposed method when a new a unseen class arrives. The black line indicates the confidence of xClass. As the fall is detected, if the inequality (\ref{eq3}) is satisfied this indicates that the label of this data sample is not any of the known to xClass labels. The options are that: a) This drop is a one off due to outlier, noise, randomness, or b) a number of such data samples above a drop of confidence is detected are close to each other in the data space (please note that they may not necessarily arrive one after the other as in Fig. \ref{fig:fall}). Otherwise, if the condition given by the inequality (\ref{eq3}) is not met the data sample is used to update the meta-parameters of the proposed method.

\begin{equation}
\begin{split}
IF~ \lambda^{max} (U_i) <   (\bar{\mu}_i - m\sigma)  ~THEN ~ (U_i \in Possible~new~class~detected)  \\
ELSE~ (Update~ structure ~and~meta-parameters)
\end{split}
\label{eq3}
\end{equation}

When the inequality (\ref{eq3}) is satisfied, the arrival data sample is denoted as a potential outlier and temporarly saved. When several of potential outliers are close to each other in the data space, have similar densities, they are denoted as `new class 1', if more than one group is formed than new classes are formed as well and new labels as `new class 2' are generated. The user can be proactively asked to (optionally) label with a semantically meaningful identification.

One or few labels for new detected classes are provided. The validation process is done through the `winners-take-all' principle, which is given by,

\begin{eqnarray}
Label = {argmax}(\lambda(\Bar{x})).
\label{eq5}
\end{eqnarray}

\section{Results}

In this section we will demonstrate the results obtained by the proposed extremely weakly supervised classification approach. Computational simulations were performed to assess the accuracy of the classification methods considering 2 different benchmark problems. The results from experimentation with the proposed algorithm aim to demonstrate that it offers:

\begin{itemize}
    \item[--] high precision as compared with the top state-of-the-art algorithms.
    \item[--] ability to detect unseen/new data patterns autonomously and learn from them.
    \item[--] ability to learn with extremely low supervision (few) labeled data samples for the newly detected classes.
    \item[--] ability to autonomously select the most effective features per class.
    \item[--] highly transparent interpretable model.
    \item[--] no user- or problem- specific algorithmic parameter (except for feature selection which can be done by \textit{ad hoc} decision).
    \item[--] non-iterative algorithm able to learn continuously.
\end{itemize}

\subsection{iRoads dataset}
\begin{figure}
       \centering
         \includegraphics[width=12cm]{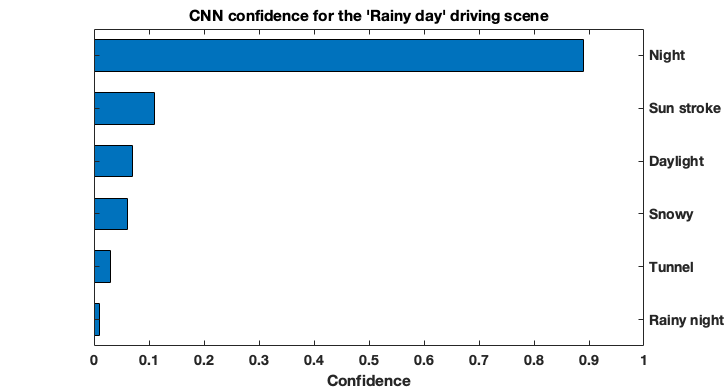}
         \caption{Wrong classification given by VGG--VD--16 for a  new unknown classes.}
         \label{fig:CNN_confidence}
\end{figure}

In the first experiment the iRoads dataset \cite{rezaei2013vehicle,8851842} was considered. 
The convolutional deep neural network VGG--VD--16 was trained with 80\% of the available iRoads dataset; however, images for the `Rainy day' scenario were omitted of the training phase. After the training phase, `Rainy day' trained images were presented to the neural network. As the VGG--VD--16 approach was not trained for the presented situation, and it is not able to adapt its structure for the newly arrived class , it misclassified the `Rainy Day' scenario with almost $90\%$ confidence as a `Night' scenario as illustrated by Fig. \ref{fig:CNN_confidence}.

\begin{figure}
         \centering
         \includegraphics[width=12cm]{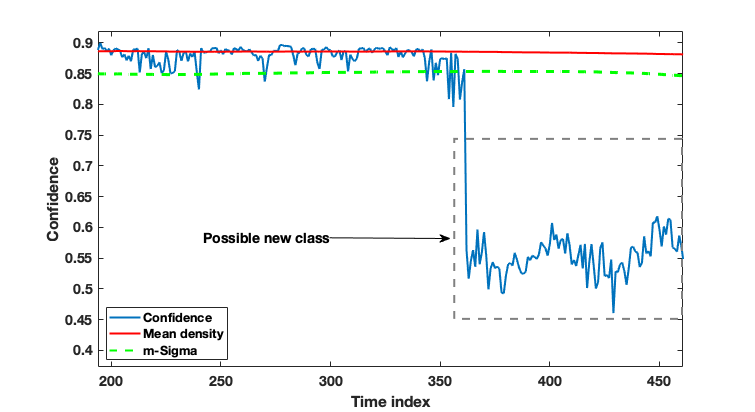}
         \caption{Sudden drop of confidence due the presentation of new unknown classes.}
         \label{fig:drop_rainy}
     \end{figure}

The convolutional neural network VGG--VD--16 misclassifed with almost 90\% of confidence the `Rainy day' driving scenario as a `Night' scenario as it was not trained for this situation as illustrated by Fig. \ref{fig:CNN_confidence}. As the VGG--VD--16 is not equipped with any exploratory mechanism it is not able to learn from these unknown data samples. Therefore, a full retraining is required for a correct classification of the presented set of images; however, a retraining of a deep neural network can be time consuming, computational expensive, and costly. The xClass exploratory mechanism is able to discover new classes as they arrive to the system due to its mechanism based on the recursive density estimation \cite{angelov2012autonomous} and Chebyshev inequality approach \cite{rousseeuw2011robust} as given by Fig. \ref{fig:drop_rainy}. The blue line indicates the confidence value ($\Lambda^{max}$ boundary) given by the xClass classifier, the orange line indicates the the recursive density estimation value, the green line is the 3-$\sigma$. The sudden fall of the blue line indicates the moment when the unlabeled set of images arrive to the system. Then, through the proposed automatic novelty detection mechanism given by the inequality (13) a new rule is created proactively if a novelty is detected. Differently from traditional deep learning approaches no retraining is required for new unseen classes as the exploratory mechanism allows the algorithm to learn from unknown.

\begin{figure}
	\begin{center}
$R_{new}$:	IF ($Image$ $\sim$ {\includegraphics[scale=0.2]{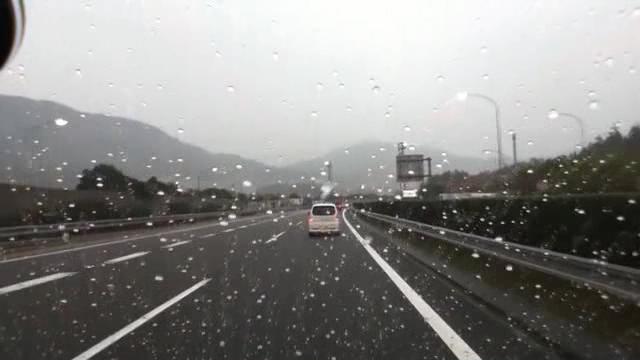} )~THEN `New class'\caption{A new rule is proactively created when a sudden fall in the confidence is detected through the inequality (13). The proposed xClass classifier is highly interpretable due to its rule-based nature. This advantage favors human experts analysis as it provides a transparent structure, differently from the `black box' approaches such as deep neural networks. } \label{rule}} 
	\end{center}
\end{figure}

The proposed xClass classifier was trained with 80\% of the available iRoads images of all classes except the `Rainy day' class. Then, the new unlabeled class was present to the proposed classifier, xClass was able to successfully detect the suddenly drastic fall in the confidence (Fig. \ref{fig:drop_rainy}) and proactively create a new class as illustrated by Fig. \ref{rule}. The prototype-based and non-iterative nature of the proposed method allowed to detect the fall in the confidence ($\lambda^{max}$) in real time, and differently, from traditional deep learning approaches, no retraining is required to learn the new class.

\begin{figure}
	\begin{center}
$R_{7}$:	IF ($Image$ $\sim$ {\includegraphics[scale=0.15]{Figure4.jpeg}) ~ OR ~ ($Image$ $\sim$ {\includegraphics[scale=0.15]{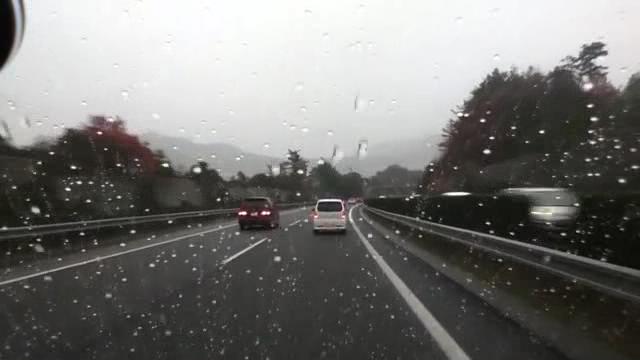}})~OR\\~... OR~ ($Image$ $\sim$ {\includegraphics[scale=0.15]{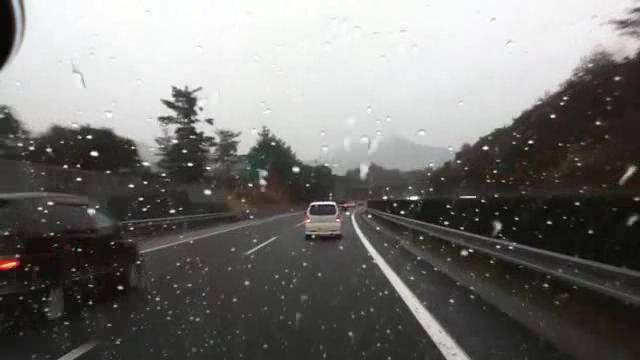}}) ~THEN ~`Rainy day scene'\caption{Final rule given by the xClass classifer for the new detected class. Label is attached during the validation phase. Differently from `black box' approaches as deep neural networks, xClass provides highly interpretable rules which can be used by human experts for different analysis as necessary.  } \label{final_rule}} 
	\end{center}
\end{figure}

The proposed xClass classifier obtained 99.12\% classification accuracy for unlabeled images using the 3-$\sigma$ approach. The semantically meaningful label `Rainy Day Scene' is optional and requires only one-off involvement by the human (by default it will stay as `new class 1'). The final rule generated for this new class detected by the proposed xClass classifier is given by Fig. \ref{final_rule}.

\subsection{Caltech-101 dataset}
\begin{figure}
    \centering
        \includegraphics[width=12cm]{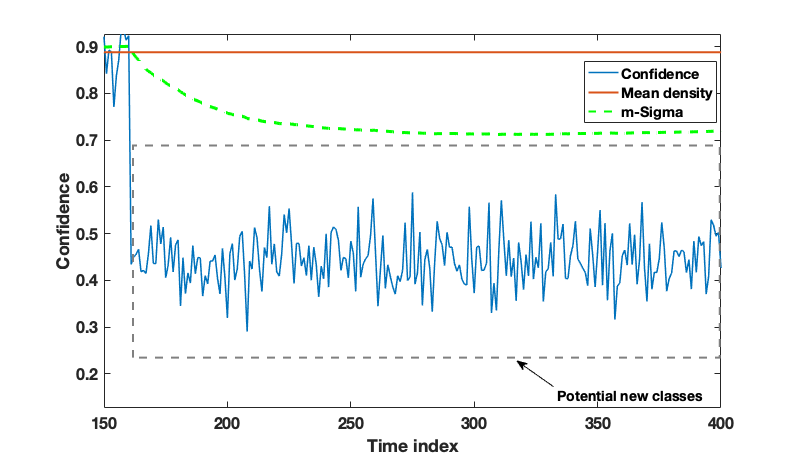}
         \caption{Number of known classes during the training phase.}
         \label{fig:shuffle_caltech}
     \end{figure}

\begin{figure}
         \centering
         \includegraphics[width=12cm]{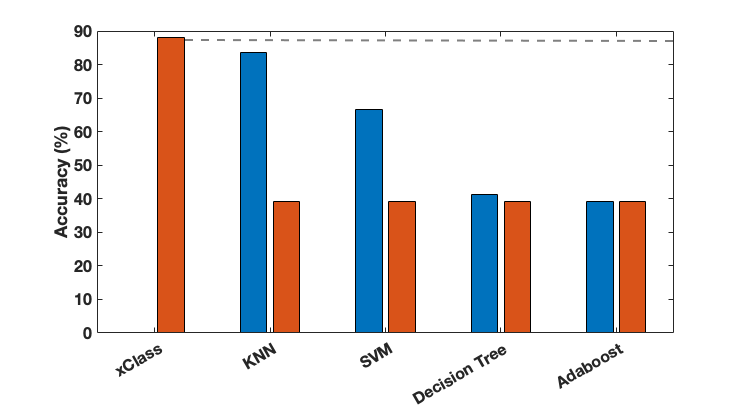}
         \caption{Accuracy for extremely weak supervision classification.}
         \label{fig:ews_caltech}
     \end{figure}

\begin{figure}
         \centering
         \includegraphics[width=12cm]{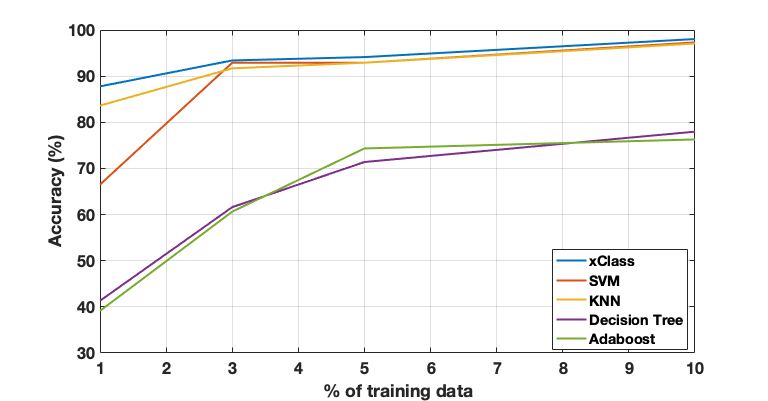}
         \caption{Classification curve for different number of training samples.}
         \label{fig:Caltech}
     \end{figure}

 As a second case, we consider the Caltech-101 dataset \cite{fei2007learning}. As in the other experiments the proposed xClass classifier was primed with 80\% of data samples from the first class for training, and then, used its exploratory mechanism to discover the other classes autonomously and learn from them based on the data density according through the drop of confidence as detailed in Fig. \ref{fig:shuffle_caltech}; as illustrated in Fig. \ref{fig:ews_caltech}, traditional approaches are not able to detect new data patterns after the training phase (traditional approaches were trained with just 1 class), and then, tend to produce results with low accuracy. Unlike supervised methods which are data hungry, the proposed xClass approach could obtain high classification accuracy with extremely weak supervision (Fig. \ref{fig:ews_caltech}), in order word, with less training data as possible. The acquisition of labeled data requires enormous human efforts and it is very time consuming. Fig. \ref{fig:Caltech} gives the evolution of the performance of the proposed exploratory classifier as more training samples are provided. As it is illustrated by Fig. \ref{fig:Caltech}, the xClass classifier is able to produce better results in terms of accuracy, demonstrating its efficiency to detect and learn from unknown effectively.   

The Caltech-101 dataset is constituted of 101 different classes. However, in the experiment only 10 classes were used. Supervised methods such as Decision tree, KNN, Adaboost, and SVM require information about all the available classes beforehand, in order to produce better results (the orange bars in Fig. \ref{fig:ews_caltech} illustrates the results obtained when just one class is used in the training phase). In comparison, the proposed extremely weakly supervised approach requires just the knowledge about one class beforehand as illustrated by Fig. \ref{fig:shuffle_caltech} as the other classes are discovered through its exploratory mechanism. The blue bar in Fig. \ref{fig:ews_caltech} illustrates the classification results when just 1\% of labeled training data is provided for all classes. The proposed exploratory xClass classifier could obtain almost 90\% of classification accuracy. State-of-the-art approached have the necessity for labeled training data to produce acceptable results as illustrated in Fig. \ref{fig:Caltech}. Even when more labeled training data is provided, the proposed xClass classifier still produce better results in terms of accuracy than its competitors.

\section{Conclusion}

In this paper, we break with the traditional approach to supervised classification. We offer a new fully autonomous extremely weakly supervised approach (xClass) which is able to learn from just a single class and a handful of labeled data samples. Then, as new classes, unknown to the human user the trained classifier appear at a later stage, the proposed xClass method is able to successfully discover this and learn from the data autonomously as demonstrated in the results section.  Results demonstrated that the proposed approach offers a high precision as compared with the top state-of-the-art algorithms. It could surpass its competitors in terms of accuracy for all experiments using extremely weak supervision, as well as, full supervision. Moreover, the proposed algorithm produced  highly transparent interpretable results, which are helpful for human experts analysis. No user- or problem- specific algorithmic parameter (except for feature selection which can be done by \textit{ad hoc} decision) are required which is also an advantage provided by the proposed xClass classifier. The proposed xClass method demonstrated he ability to learn from unknown without retraining, which is one of the biggest problems of deep learning based on neural networks. As illustrated the convolutional deep learning misclassified an unknown class with high confidence, in the other hand, the proposed approach was able to detect a sudden drop in the confidence and learn from this unknown data, then it was able to proactively create a new class for this new scenario. The proposed method is applicable to a wide range of problems, especially for problems with unknown dimension and for problems for which the concept changes over time.
\bibliographystyle{unsrt}  
\bibliography{references}  


\end{document}